\documentclass[dvipsnames,format=sigconf,anonymous=false,review=false, nonacm=True]{acmart}

\AtBeginDocument{%
  \providecommand\BibTeX{{%
    \normalfont B\kern-0.5em{\scshape i\kern-0.25em b}\kern-0.8em\TeX}}}

\usepackage{multirow}
\usepackage{algorithm}
\usepackage{algorithmic}
\usepackage{caption}
\usepackage{graphicx}
\usepackage{amsmath}

\usepackage{amsthm}
\usepackage{amsfonts,amssymb,mathrsfs}
\usepackage{soul}
\usepackage{subfigure}
\usepackage{mathtools}
\usepackage{float}
\usepackage{verbatim}
\usepackage{booktabs}
\usepackage{svg}
\usepackage{enumitem}
\usepackage{xcolor}

\copyrightyear{2024}
\acmYear{2024}
\setcopyright{acmlicensed}\acmConference[GECCO '24]{Genetic and Evolutionary Computation Conference}{July 14--18, 2024}{Melbourne, VIC, Australia}
\acmBooktitle{Genetic and Evolutionary Computation Conference (GECCO '24), July 14--18, 2024, Melbourne, VIC, Australia}
\acmDOI{10.1145/3638529.3653995}
\acmISBN{979-8-4007-0494-9/24/07}

\begin{document}

\title{RLEMMO: Evolutionary Multimodal Optimization Assisted \\By Deep Reinforcement Learning}
\renewcommand{\shorttitle}{RLEMMO: Evolutionary Multimodal Optimization Assisted By Deep Reinforcement}

\author{Hongqiao Lian}
\email{qiao2471035068@163.com}
\orcid{0009-0005-4424-6189}
\affiliation{%
  \institution{South China University of Technology}
  \city{Guangzhou}
  \state{Guangdong}
  \country{China}
}

\author{Zeyuan Ma}
\authornote{Corresponding Author}
\email{scut.crazynicolas@gmail.com}
\orcid{0000-0001-6216-9379}
\affiliation{%
  \institution{South China University of Technology}
  \city{Guangzhou}
  \state{Guangdong}
  \country{China}
}

\author{Hongshu Guo}
\email{guohongshu369@gmail.com}
\orcid{0000-0001-8063-8984}
\affiliation{%
  \institution{South China University of Technology}
  \city{Guangzhou}
  \state{Guangdong}
  \country{China}
}

\author{Ting Huang}
\email{gnauhgnith@gmail.com}
\orcid{0000-0002-8755-043X}
\affiliation{%
  \institution{Xidian University}
  \city{Xi'an}
  \state{Shanxi}
  \country{China}
}

\author{Yue-Jiao Gong}
\email{gongyuejiao@gmail.com}
\orcid{0000-0002-5648-1160}
\affiliation{%
  \institution{South China University of Technology}
  \city{Guangzhou}
  \state{Guangdong}
  \country{China}}

\renewcommand{\shortauthors}{Lian et al.}

\begin{abstract}
  Solving multimodal optimization problems (MMOP) requires finding all optimal solutions, which is challenging in limited function evaluations. Although existing works strike the balance of exploration and exploitation through hand-crafted adaptive strategies, they require certain expert knowledge, hence inflexible to deal with MMOP with different properties. In this paper, we propose RLEMMO, a Meta-Black-Box Optimization framework, which maintains a population of solutions and incorporates a reinforcement learning agent for flexibly adjusting individual-level searching strategies to match the up-to-date optimization status, hence boosting the search performance on MMOP. Concretely, we encode landscape properties and evolution path information into each individual and then leverage attention networks to advance population information sharing. With a novel reward mechanism that encourages both quality and diversity, RLEMMO can be effectively trained using a policy gradient algorithm. The experimental results on the CEC2013 MMOP benchmark underscore the competitive optimization performance of RLEMMO against several strong baselines. 
\end{abstract}





\maketitle

\section{Introduction}
Multimodal optimization problems (MMOPs) involve multiple optimal solutions, which are frequently encountered in real-world tasks such as truss structure design~\cite{truss-structures1, truss-structures2}, job scheduling~\cite{jobscheduling}, electromagnetic machine design~\cite{electromagnetic}, etc. One of the challenging tasks in solving MMOPs is to locate as more as possible optimal solutions within a limited number of function evaluations, which has been extensively discussed in recent literature.

Evolutionary algorithms~(EAs) have demonstrated robustness and effectiveness in incorporating meta-heuristics to efficiently locate global optima~\cite{DE, pso}. However, classic EAs face challenges when addressing MMOPs due to diversity loss caused by inherent genetic drift and selection pressure. To overcome these bottlenecks, a significant amount of research has focused on enhancing classic EAs to solve MMOPs. Three major branches of study are widely discussed in the literature: 1) Incorporating niching methods~\cite{VNCDE, LIPS} with classic EAs to reinforce selection flexibility~\cite{CDE} and maintain diversity~\cite{sharing, species} within the population.
2) Exploiting historical search experiences to assist population reproduction through informative sub-space sampling~\cite{EBSP}, reproduction redundancy checking~\cite{darchive}, and optimum backtracking~\cite{hisarchive}. 3) Transforming MMOPs into multi-objective optimization problems~\cite{MOMMOP, EMOMMO} by introducing two or more conflicting objective functions to balance solution diversity and quality. While these enhancements enable classic EAs to adaptively coordinate exploitation and exploration by balancing quality and diversity during the optimization process, they often require expert knowledge to hand-craft these adaptive mechanisms. This can make the approach labor-intensive and inflexible when dealing with MMOPs with different properties.

Recently, works in Meta-Black-Box Optimization~(MetaBBO) have successfully utilized neural networks to meta-learn configurations of traditional black-box optimizers, thereby reducing the reliance on deep expert knowledge in existing MMOP algorithms~\cite{NEURIPS2023_232eee8e}. However, most existing MetaBBO works are developed for other optimization scenarios, and only a few have been specifically designed for addressing MMOPs. One recent work in this domain is RLEA-SSC~\cite{rlmmo}, which meta-learns an adaptive selection pressure mechanism to enhance population diversity as the evolution progresses. However, RLEA-SSC requires re-training for each unseen problem instance, significantly limiting its generalization ability. To develop a generalizable MetaBBO method tailored for MMOPs, in this paper, we propose \textbf{\ul{E}}volutionary \textbf{\ul{M}}ulti\textbf{\ul{m}}odal \textbf{\ul{O}}ptimization Assisted by Deep \textbf{\ul{R}}einforcement \textbf{\ul{L}}earning (RLEMMO), an MetaBBO framework that incorporates an RL agent at the meta-level to flexibly control the search behaviors of the population at the lower level optimization process. The RL agent is trained at the meta level to maximize both the quality and diversity in the low-level optimization process, effectively addressing the challenges in MMOPs. Specifically, we construct a comprehensive state representation based on fitness landscape analysis~\cite{fla} to capture quality and diversity information at both the population and individual levels during the optimization process. We develop an attention-based network structure for efficient feature embedding and search behavior control. Additionally, a clustering-based reward scheme is proposed to effectively meta-train the RL agent and enhance optimization performance. We train RLEMMO using policy gradient methods on a group of MMOPs and then generalize the trained model to unseen problem instances. Experimental results demonstrate that our RLEMMO achieves competitive optimization performance against several strong baselines which are tailored for MMOPs.

We now summarize the contributions in this paper:
\begin{itemize}
    \item Introduction of RLEMMO, a pioneering MetaBBO framework that effectively solve MMOPs through a meta-learned policy, which enriches the quality and diversity of the lower level optimization process.
    
    \item A comprehensive state representation based on fitness landscape analysis that helps RLEMMO generalize to unseen problems, along with a novel clustering-based reward scheme that assists the RL agent in learning effective policies.
    
    \item We demonstrate the effectiveness of RLEMMO on well-known MMOP benchmark problems. The results show that RLEMMO is competitive with several strong methods. Furthermore, we provide ablation studies that give in-depth analysis on the core designs in RLEMMO.
\end{itemize}

\section{Related Works}

\subsection{Evolutionary Multimodal Optimization}\label{Sec2.1}
Several works have made significant contributions for solving MMOPs by enhancing evolutionary computation from various perspectives.

The first aspect of the improvement involves integrating Evolutionary Computation (EC) with niching methods to enhance the diversity of the population during optimization. Thomsen et al.~\cite{CDE} proposed Crowding DE, which combines differential evolution with the crowding method. Each offspring replaces the nearest individuals if it has a better fitness. Qu et al. \cite{LIPS} introduced a locally informed particle swarm optimizer to enhance niching PSO by combining several local best solutions in the vicinity of each particle to guide the velocity update. Li et al.~\cite{ring} proposed a ring topology PSO that eliminates the need for an extra parameter. The swarm is divided using a ring topology. Zhang et al. \cite{proximity} proposed a method for choosing parents based on proximity ranking, where the probabilities of selecting neighbors as parents are determined by their distance from the individual. To adaptively adjust niching radii to align with the dynamical optimization process, Zhang et al. \cite{VNCDE} proposed VNCDE, which first constructed a Voronoi neighborhood for each individual and thereby adaptively assigned desired searching strategies within the neighborhood. Luo et al.~\cite{NBNC} proposed an adaptive nearest-better neighbor clustering method for characterizing basins of attraction, which not only significantly enhances the diversity during the optimization process, but also provides flexible convergence.
 
The second aspect of the improvement leans utilizing an archive to assist optimization with historical experience. In particular, The dADE method~\cite{darchive} identifies redundant individuals for reinitialization using an archive that stores promising solutions. Huang et al. \cite{EBSP} proposed storing historical information in an enhanced binary space partitioning tree, dividing the space with the tree to aid optimization. Liao et al. \cite{hisarchive} developed a history archive-assisted niching differential evolution with variable neighborhood, using the information from the archive to control the size of the neighborhood and change it during the evolution process. Wang et al. \cite{Aestimation} proposed a parameter-free niching method based on adaptive estimation distribution, where each individual estimates the distribution of the environment based on nearby neighbors and decides the size of the niching.

As for the third aspect of the improvement, several recent works solve MMOPs through transforming them into multi-objective optimization problems~(MOOs). These works commonly considered two objectives: the solution quality and the solution diversity of the population~\cite{MMO1, MMO2}. Wang et al. \cite{MOMMOP} highlighted the significance of conflicts between the two objectives and proposed a novel method to efficiently construct conflict objectives for problem transformation. Cheng et al. \cite{EMOMMO} transformed MMOPs into multi-objective problems using multi-objective EAs to obtain an approximate fitness landscape and then detected peaks based on stored historical samples. A local search method was applied to optimize the global optima.

Though promising results have been achieved by the above evolutionary multimodal optimization methods, they often require intricate manual tuning with deep expertise to secure the desirable performance for unseen tasks. The labour-intensive design process of these human-crafted methods hinders both the efficiency and effectiveness of solving MMOPs.

\subsection{Meta-Black-Box Optimization}
Recent Meta-Black-Box Optimization (MetaBBO) approaches~\cite{NEURIPS2023_232eee8e, chen2024symbol, guo2024deep},  facilitate a bi-level optimization scheme to meta-learn a parameterized control policy in a data-driven way, which configures the low-level black-box optimizer during the optimization process to mitigate the design efforts required for unseen tasks.The parameterized control policy at the meta level can be optimized in various ways~(e.g., supervised learning~\cite{chenLearning, tv2019meta, gomes2021meta}, reinforcement learning~\cite{Sharma, tan2021differential, Sun}, and self-referential searching~\cite{gomes2021meta, lange2023discovering, lange2023discovering2}). Take MetaBBO with reinforcement learning~(MetaBBO-RL) as an example, it models the fine-tuning of the optimizer as a Markov Decision Process (MDP) and trains an agent to make decisions regarding algorithmic configurations at the meta-level. At each step $t$, the agent queries the optimization status $s_t$, and the policy $\Pi_\theta$ uses $s_t$ as input to suggest an action $a_t$ for configuring the optimizer at the lower level. The environment then executes $a_t$ to obtain a new status $s_{t+1}$. A reward $r_t$ is generated to measure the meta-performance improvement of the optimizer at the lower level. Subsequently, the agent at the meta-level can be trained based on the rules of reinforcement learning. The action space of MetaBBO-RL can involve dynamically selecting evolutionary algorithms/operators~\cite{guo2024deep, Sharma, tan2021differential}, controlling hyper-parameters~\cite{Sun, wu2022employing}, or generating interpretable update rules~\cite{chen2024symbol}. This paper shares the similar ambition with the MetaBBO-RL for selecting algorithms/operators. Along this research branch, Sharma et al.~\cite{Sharma} trained a Deep Q-Network~(DQN)~\cite{dqn} to dynamically select the mutation strategy for the backbone DE optimizer, for each individual based on the timely optimization status. Guo et al.~\cite{guo2024deep}, on the other hand, propose leveraging Proximal Policy Optimization~(PPO)~\cite{ppo} to train an RL agent which selects suitable optimizer to optimize the problem for different optimization phases, enhancing the final optimization performance. To the best of our knowledge,  in the domain of using MetaBBO-RL to solve MMOPs, Xia et al.~\cite{rlmmo} proposed to use Q-Learning to select solutions within the approximated basin of attraction, which balances the quality and diversity of the solutions during the optimization. However, this method needs re-training for unseen problems, lacking of the ability of generalization. 


\section{Preliminary}

\subsection{K-nearest Neighbors(KNN)}

KNN \cite{KNN} identifies the class of point $X_i$ based on its $k$ neighbors. Firstly, the Euclidean distance between $X_i$ and other points is calculated as a measure of similarity. The $k$ points with the shortest distances are considered the k-nearest neighbors of $X_i$, and the class of $X_i$ is determined based on its neighbors. In this paper, the KNN method is used to construct the k-nearest neighborhood for each individual. Then, the features about the neighborhoods are calculated to provide local information to each individual, as described in Section~\ref{Sec4.2}.
\begin{figure}[t]
  \centering
  \includegraphics[width=0.6\columnwidth]{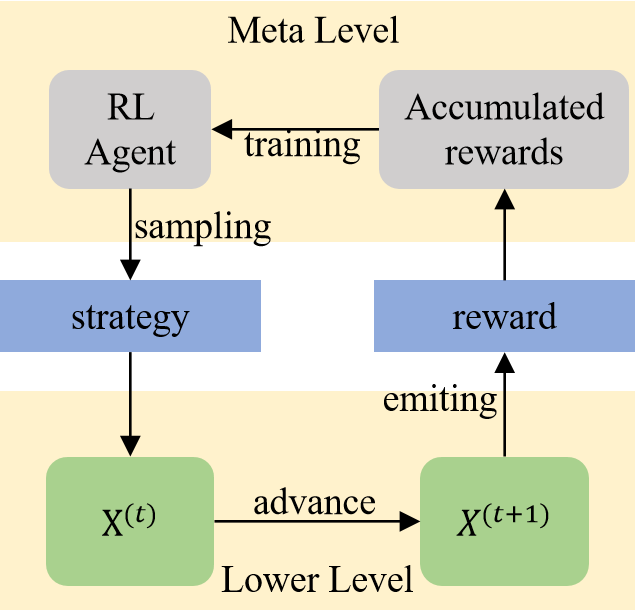}
\caption{Blueprint of RLEMMO, where the meta-level RL agent outputs a search strategy for advancing the solution population in the low-level optimization. The meta-level RL agent is meta-trained to maximize the accumulated reward during the low-level optimization.}
  \label{fig:blurprint}
\end{figure}
\subsection{DBSCAN}

Density-Based Spatial Clustering of Applications with Noise (DBSCAN) \cite{DBSCAN} is an unsupervised clustering algorithm. It is a non-parametric method based on density. For a set of points, the points that are closely packed together are grouped as clusters, while the points that lie alone in low-density regions are regarded as noise points. In this paper, RLEMMO utilizes DBSCAN to cluster the population in each generation, estimating the number of clusters and the members within each cluster. A novel reward mechanism based on the number of clusters and the best fitness within each cluster is designed to encourage the diversity and quality of the population during the optimization, as described in detail in Section~\ref{Sec4.4}.

\subsection{Policy Gradients}

Policy gradients is a method used to refine the Markov Decision Process (MDP) \cite{RL}. Given an MDP with the form $<S, A, \mathcal{T}, R>$, the agent observes a state $s \in S$ and selects an action $a \in A$. The state $s$ transitions to $s'$ according to the dynamics of the environment $\mathcal{T}(s'|s, a)$, and the agent receives a reward $R(s,a)$ from the environment. The objective of RL is to find a policy $\pi_\Theta(a|s)$, parameterized by the model weights $\Theta$, that maximizes the cumulative discounted reward $E_{\pi_\Theta}(\sum_{t=1}^T \gamma^{t-1}R(s^{(t)}, a^{(t)}))$, denoted as $J(\Theta)$. Policy gradients aim to maximize the expected value through gradient ascent. The gradient of $J(\Theta)$ with respect to $\Theta$, denoted as $\nabla J(\Theta)$, can be calculated as follows:

\begin{equation}
\nabla J(\Theta) = E[R^{(t)} \nabla{\log{\pi_{\Theta}(a|s)}}]
\end{equation}
where $R^{(t)}$ represents the cumulative reward at time $t$. Due to the effectiveness of policy gradients, various extensions have been developed, such as REINFORCE \cite{REINFORCE}, TRPO \cite{TRPO}, and PPO \cite{ppo}. In this paper, we utilize PPO to refine the MDP of MMOPs.

\subsection{Attention Mechanism}

The attention mechanism \cite{attention} is calculated as follows:

\begin{equation}
Attn(Q,K,V) = \text{softmax}\left(\frac{QK^\text{T}}{\sqrt{d_k}}\right)V
\end{equation}
where $Q$, $K$, and $V$ are vectors representing the query, key, and value, respectively. These vectors are computed by weighting the input sequence with $W_q$, $W_k$, and $W_v$. The value of $d_k$ represents the dimension of the key vector and is used for result normalization. In the case of the self-attention mechanism, $Q$, $K$, and $V$ are derived from the same source sequence. 

The multi-head attention mechanism maps vectors to different subspaces to enhance representation. Each attention head independently calculates $Q_i$, $K_i$, and $V_i$. The outputs of all heads are then concatenated together to produce the final result of the multi-head attention module:

\begin{equation}
\begin{aligned}
H = \text{MHA}(Q,K,V) = \text{Concat}(H_1,H_2,...,H_h)W_o\\
H_i = \text{Attn}(Q_i, K_i, V_i)
\end{aligned}
\end{equation}
In our work, we utilize an attention module to enhance information sharing among the individuals in RLEMMO.

\begin{table*}[t]
  \caption{Individual-level state formulations and detailed descriptions.}
  \label{tab:state}
  \resizebox{0.9\textwidth}{!}{%
  \begin{tabular}{cccc}
    \toprule
    Class &Number &Formula  & Description\\
    \midrule
    \multirow{11}{*}{\rotatebox{0}{$f_{g}$}}
    &1
    &$\text{mean}_{i,j}(\frac{\| X_i^{(t)} - X_j^{(t)} \|_2}{diameter})$ 
    &\begin{tabular}[c]{@{}c@{}}
    The average distance between each pair in the population at generation $t$\\ normalized by the diameter of the searching space.
    \end{tabular}\\
    \cline{2-4} 
    &2
    & $ \text{std}_i(\frac{Obj(X_i^{(t)})}{Obj_\text{max}}) $ 
    &\begin{tabular}[c]{@{}c@{}} 
    The standard deviation of individual objective values at generation $t$.\\$Obj(X_i^{(t)})$ denotes the objective value of $X_i^{(t)}$. \\Normalized by the maximal objective value gap $Obj_\text{max}$.\\$Obj_{max}$ is defined as the  gap between the historical worst  and the historical best objective value in an episode.
     \end{tabular}\\
    \cline{2-4} 
    &3& $\frac{(T-t)}{T} $ & \begin{tabular}[c]{@{}c@{}}
    The remaining portion of generations at generation $t$. \\Normalized by the  maximal number of generations $T$.   \\
     \end{tabular}\\
    \cline{2-4} 
    &4& $\frac{st}{T}$& \begin{tabular}[c]{@{}c@{}}The stagnation of the population. \\ $st$ represents the number of generations since the last update of the global best object value.\\Normalized by $T$.\\ \end{tabular}\\
    \cline{2-4} 
    &5& $ \text{mean}_i(\frac{Obj(X_i^{(t)})}{Obj_\text{max}}) $& \begin{tabular}[c]{@{}c@{}}The average objective value of all individuals in the population at generation $t$. \\Normalize by the maximal objective value gap $Obj_\text{max}$. \end{tabular}\\
    \cline{1-4} 
    \multirow{10}{*}{\rotatebox{0}{$f_{n}$}}&6& $\text{mean}_{j,k\in \Phi_i^{(t)}}(\frac{{\|X_j^{(t)} - X_k^{(t)}\|}_2}{diameter})$& \begin{tabular}[c]{@{}c@{}}The average distance between each pair of individuals within the neighborhood of $X_i^{(t)}$, denoted as $\Phi_i^{(t)}$. \\Normalized by the diameter in the space. \end{tabular}\\
    \cline{2-4} 
    &7& $\text{std}_{j\in \Phi_i^{(t)}}(\frac{Obj(X_j^{(t)})}{Obj_\text{max}})$& \begin{tabular}[c]{@{}c@{}}The standard deviation of objective value within the neighborhood $\Phi_i^{(t)}$. \\Normalized by  $Obj_\text{max}$.\end{tabular}\\
    \cline{2-4} 
    &8& $\frac{st(\Phi_i)}{T} $& \begin{tabular}[c]{@{}c@{}}Stagnation within $\Phi_i$ , where $st(\Phi_i)$ denotes the number of generation since last update of the best value within $\Phi_i$. \\Normalized by the maximal number of generation $T$. \end{tabular}\\
    \cline{2-4} 
    &9 &$\text{mean}_{j\in \Phi_i^{(t)}}(\frac{Obj(X_j^{(t)})}{Obj_\text{max}})$&  \begin{tabular}[c]{@{}c@{}}The average objective value within the neighborhood $\Phi_i^{(t)}$. \\Normalized by  $Obj_\text{max}$.\end{tabular}\\
    \cline{2-4} 
    &10& $\frac{\text{rank}({\Phi_i^{(t)}})}{(NP-1)}$&   \begin{tabular}[c]{@{}c@{}}The rank of the neighborhood $\Phi_i^{(t)}$ among all neighborhoods in the the population at generation $t$.\\ Normalized by the maximal rank($NP$ -1), where $NP$ is the size of the population.\end{tabular}\\
    \cline{1-4} 
    \multirow{23}{*}{\rotatebox{0}{$f_{\text{ind}}$}}&11& $\frac{{\|X_i^{(t)} - X^{*,{(t)}}\|}_2}{diameter}$&  \begin{tabular}[c]{@{}c@{}}The distance between $X_i$ and the best solution in the population at generation $t$, denoted as $X^{*,{(t)}}$.\\ Normalized by $diameter$.\end{tabular}\\
    \cline{2-4} 
    &12& $\frac{{\|X_i^{(t)} - X^*\|}_2}{diameter}$&  \begin{tabular}[c]{@{}c@{}}The distance between $X_i^{(t)}$ and the historical best solution $X^*$. \\Normalized by $diameter$.\end{tabular} \\
    \cline{2-4} 
    &13& $\frac{Obj(X_i^{(t)}) - Obj(X^*)}{Obj_\text{max}}$&\begin{tabular}[c]{@{}c@{}} The objective value gap between $X_i^{(t)}$ and  $X^*$. \\Normalized by $Obj_\text{max}$.\end{tabular}\\
    \cline{2-4} 
    &14& $\frac{Obj(X_i^{(t)}) - Obj(X^{*,{(t)}})}{Obj_\text{max}}$& \begin{tabular}[c]{@{}c@{}}The objective value gap between $X_i^{(t)}$ and $X^{*,{(t)}}$.\\ Normalized by $Obj_\text{max}$.\end{tabular}\\
    \cline{2-4} 
    &15& $\frac{{\|X_i^{(t)} - \Phi_i^{*}\|}_2}{diameter}$&  \begin{tabular}[c]{@{}c@{}}The distance between $X_i^{(t)}$ and the best solution in $\Phi_i^{(t)}$ at generation $t$, denoted as $\Phi_i^{*}$.\\Normalized by $diameter$.\end{tabular}\\
    \cline{2-4} 
    &16& $\frac{Obj(X_i^{(t)}) - Obj(\Phi_i^{*})}{Obj_\text{max}}$& \begin{tabular}[c]{@{}c@{}}The objective value gap between $X_i^{(t)}$ and $\Phi_i^{*}$. \\Normalized by $Obj_\text{max}$.\end{tabular}\\
    \cline{2-4} 
    &17& $\frac{st(i)}{T}$&  \begin{tabular}[c]{@{}c@{}}The stagnation of $X_i$ , where $st(i)$ denotes the number of generations since last update of $X_i$ . \\Normalized by $T$.\end{tabular}\\
    \cline{2-4} 
    &18& $\frac{Obj(X_i^{(t)})}{Obj_\text{max}}$& \begin{tabular}[c]{@{}c@{}}The objective value of $X_i$ at generation $t$. \\Normalized by $Obj_\text{max}$.\end{tabular}\\
    \cline{2-4} 
    &19& $\text{mean}_{j \in \Phi_i^{(t)}}(\frac{Obj(X_i^{(t)}) - Obj(X_j^{(t)})}{Obj_\text{max}})$&  \begin{tabular}[c]{@{}c@{}}The average objective value gap between $X_i^{(t)}$ and its neighbors in $\Phi_i^{(t)}$. \\Normalize by $Obj_\text{max}$.\end{tabular}\\
    \cline{2-4} 
    &20& $\text{mean}_{j \in \Phi_i^{(t)}}(\frac{{\|X_i^{(t)} - X_j^{(t)}\|}_2}{diameter})$&  \begin{tabular}[c]{@{}c@{}}The average distance between $X_i^{(t)}$ and its neighbors in $\Phi_i^{(t)}$. \\Normalized by $diameter$.\end{tabular}\\
    \cline{2-4} 
    &21& $\text{mean}_j(\frac{Obj(X_i^{(t)})- Obj(X_j^{(t)})}{Obj_\text{max}})$&   \begin{tabular}[c]{@{}c@{}}The average objective value gap between $X_i$ and other individuals in the population at generation $t$. \\Normalized by $Obj_\text{max}$.\end{tabular}\\
     \cline{2-4} 
    &22& $\text{mean}_j(\frac{{\|X_i^{(t)} - X_j^{(t)}\|}_2}{diameter})$& \begin{tabular}[c]{@{}c@{}}The average distance between $X_i$ and other individuals in the population at generation $t$.\\Normalized by $diameter$.\end{tabular}\\
    \bottomrule
  \end{tabular}
  }
\end{table*}
\section{Methodology}

\subsection{Overview}
Our RLEMMO is a Meta-Black-Box Optimization framework with Reinforcement Learning (MetaBBO-RL), which follows a bi-level optimization approach. For a given MMOP, the bi-level optimization is carried out as follows: At the lower level MDP, starting with an initial solutions population $X^{(0)}$, an RL agent (a policy network) dynamically switches diverse searching strategies for each individual in the population at each time step $t$ along the lower level optimization process. The solutions population is advanced from $X^{(t)}$ to $X^{(t+1)}$ using the strategy sampled from the RL agent, and emits a reward back to the RL agent. At the meta level, the RL agent is trained to maximize the expectation of the accumulated reward (meta objective) over a MMOP distribution. The blueprint of RLEMMO is illustrated in Figure~\ref{fig:blurprint}. The components of the low-level MDP, including the optimization state representation, the solution-level optimization strategies (action), and the reward design, are detailed in Section~\ref{Sec4.2}, \ref{Sec4.3}, and \ref{Sec4.4} respectively. A step-by-step derivation of the RLEMMO's workflow is provided in Section~\ref{Sec4.5}.



\subsection{State Representation}\label{Sec4.2}
In RLEMMO, we incorporate Fitness Landscape Analysis~(FLA) \cite{fla} and Exploratory Landscape Analysis~(ELA) \cite{ELA} to profile the optimization status of the solution population during the optimization process, which are commonly adopted in recent MetaBBO approaches~\cite{Sharma, tv2019meta}. In specific, we utilize an individual-level feature $f_{\text{ind}}$ to analyze the status of each individual, as well as an additional neighbor-aware population feature $f_{\rm{pop}}=\{f_g,f_n\}$, consisting of two parts: $f_g$ profiles the distributional features considering the entire population, and $f_n$ describes the local information around each individual. Feature $f_{\text{pop}}$ complements $f_{\text{ind}}$ and helps inform the meta-level RL to coordinate exploration and exploitation in the lower-level optimization. Note that we also inject the evolution-path information into the optimization status, such as stagnation and historical best solutions, to reflect the experience of the evolutionary history. The detailed formulas and descriptions of the state representation are presented in Table~\ref{tab:state}. We leave the per-feature exegesis at Appendix~\ref{appendixA}.
\begin{figure*}[t]
  \centering
  \includegraphics[width=0.8\textwidth]{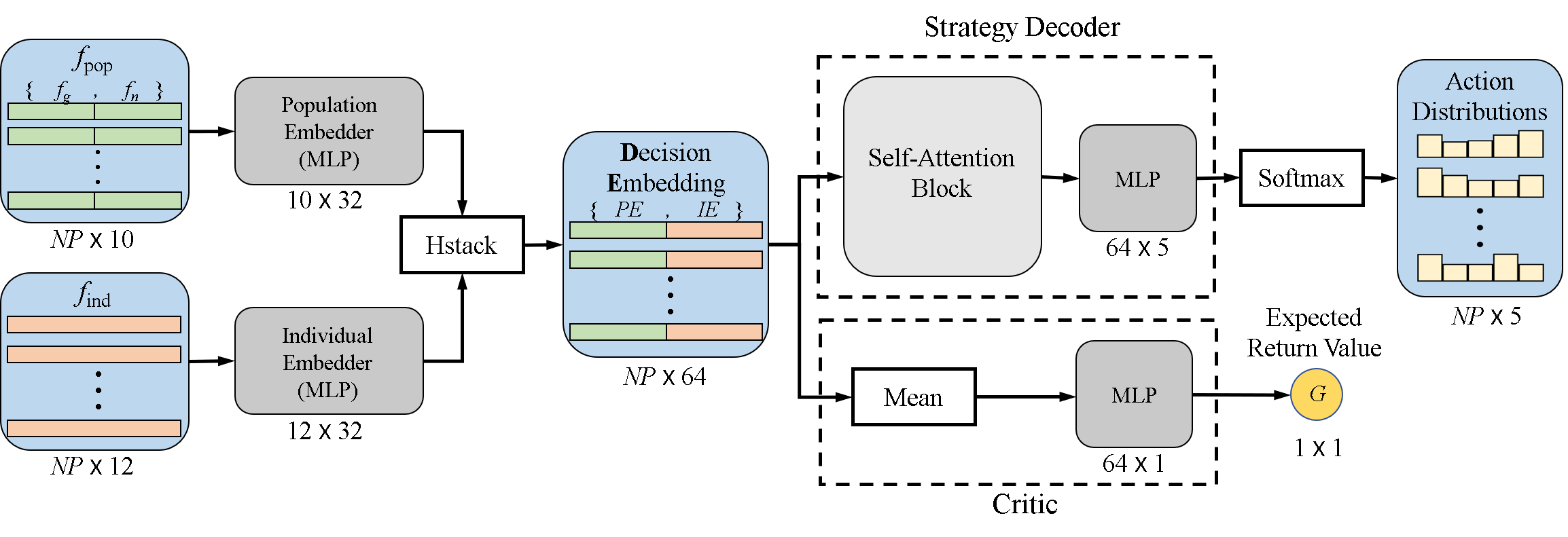}
  \caption{The architecture of neural networks in RLEMMO is depicted, with arrows indicating the overall workflow: at each time step, we input the state representation of the current solution population into the neural networks and sample individual-level strategies to advance the population along the low-level optimization process. And the critic is used to estimate the return value for training the policy network.}
  \label{fig:architecture}
\end{figure*}
\subsection{Action}\label{Sec4.3}
We provide five optional searching strategies that vary in the extent of exploration and exploitation to enrich the behavioral diversity of the low-level optimization process. Given the optimization state $f_\text{pop}$ and $f_\text{ind}$ of the individuals in the current solution population, we allow the RL agent to choose the appropriate strategies for each individual to match the up-to-date optimization status. We use $A$ to denote the action (selected strategies) for all individuals, and $A_1 \sim A_5$ to denote the five strategies respectively. These strategies are briefly introduced as follows:
\begin{enumerate}
     
\item[\textbf{A$_1$:}] Inspired by VNCDE \cite{VNCDE}, the first strategy is the Gaussian-based local search strategy, which perturbs the individual to search for a local optimum in a narrow range. Its formulation for the $i$-th individual $X_i$ is shown below:
\begin{equation}\label{eq:act1}
U_i = X_i + G(0, \sigma)
\end{equation}
where $G(0, \sigma)$ represents a random vector with values sampled from a Gaussian distribution with a mean of 0 and a standard deviation of $\sigma$. The setting for $\sigma$ follows the original VNCDE paper. $A_1$ can aid the convergence of the solution population when the solutions are approaching the global optima.
\\

\item[\textbf{A$_2$:}] We propose a KNN-based neighborhood mutation strategy to enhance the exploitation of local optimization information. It is formulated as follows:
\begin{equation}\label{eq:act2}
U_{i} = X_{i} + F \cdot (\Phi_i^{*} - X_{i}) + F \cdot (\Phi_{i, r_1} - \Phi_{i, r_2})
\end{equation}
where $\Phi_i^{*}$ represents the best individual in the KNN-based neighborhood of $X_i$(denoted as $\Phi_i$), while $\Phi_{i, r_1}$ and $\Phi_{i, r_2}$ are two distinct random individuals selected from $\Phi_i$. $F$ denotes the scaling factor.
\\

\item[\textbf{$\text{A}_3$:}] Additionally, we propose a KNN-based neighborhood mutation strategy to enhance the exploration of local optimization information. It is formulated as follows:
\begin{equation}\label{eq:act3}
U_{i} = \Phi_{i, r_1} + F \cdot (\Phi_{i, r_2} - \Phi_{i, r_3})
\end{equation}
where $\Phi_{i, \cdot}$ represents distinct random individuals selected from $\Phi_i$. $A_3$ serves as a complement to $A_2$ to ensure comprehensive exploration within the local neighborhood area.
\\

\item[\textbf{$\text{A}_4$:}] We propose a global exploratory mutation strategy that reinforces information sharing across different KNN-based neighborhoods.
\begin{equation}\label{eq:act4}
    U_{i} = X_{i} + F \cdot (\Phi_j^{*} - X_{i}) + F \cdot (X_{r_1} - X_{r_2})
\end{equation}
Using the best individual in a randomly chosen neighborhood $\Phi_j$ as $\Phi_j^{*}$, the strategy allows $X_i$ to jump out of the current area and chase another optimum. The inclusion of two randomly selected individuals, $X_{r_1}$ and $X_{r_2}$, makes the process highly exploratory.
\\

\item[\textbf{$\text{A}_5$:}] We adopt the well-known "DE/rand/1" mutation strategy to thoroughly explore the entire search space.
\begin{equation}
    U_{i} = X_{r_1} + F \cdot (X_{r_2} - X_{r_3})
\end{equation}
where $X_{r\cdot}$ represents randomly selected individuals from the population. We note that $A_4$ and $A_5$ facilitate an individual's ability to escape from both local optima and locally best neighbors.

\end{enumerate}
The aforementioned searching strategies exhibit diverse behaviors when applied to the individuals in the solution population. Considering that in RLEMMO, the search behaviour of the population is controlled by the RL agent in a fine-grained way, e.g., at the individual level. This presents an opportunity for RLEMMO to enhance low-level optimization performance through meta-learning a flexible policy.

\subsection{Reward Design}\label{Sec4.4}
In the case of MMOP, providing feedback to the meta-level RL agent during the low-level optimization process poses a challenge. In RLEMMO, the reward feedback should not only inform the RL agent about the potential optimization performance gain achieved by adopting a specific searching strategy for the next optimization step (e.g., DE-DDQN \cite{Sharma}, LDE \cite{Sun}), but it should also make the RL agent aware of the diversity within the current solution population, thereby facilitating the discovery of more global optima later on. To address this, we propose a novel reward scheme, denoted as $R_\text{clb}$, which synthetically reflects the potential performance gain in both solution quality and diversity based on clustering estimation. Specifically, when the solution population $X$ advances from time $t$ to $t+1$, we first cluster the solutions in $X^{(t+1)}$ using the DBSCAN \cite{DBSCAN} method. Let ${C_1, C_2, ..., C_k}$ represent the resulting $k$ clusters. We then evaluate the solutions in each cluster and denote the objective value of the best-performing solution in the $j$-th cluster as $f^*_j$. The clustering-based reward at time step $t$ is calculated as $R_\text{clb}^{(t)}(X^{(t)},A^{(t)})=\sum_{j=1}^k f^*_j$. The $R_\text{clb}$ strikes a balance between the quality and diversity of solutions in a way that if the meta-level RL agent in RLEMMO seeks higher rewards throughout the MDP episode, it should learn a flexible policy that efficiently locates all global optima as quickly and comprehensively as possible. We note that the proposed clustering-based reward scheme aids the efficient convergence of the meta-level RL agent while enhancing the population diversity within the low-level optimization process.

\subsection{Workflow of RLEMMO}\label{Sec4.5}
We will now outline the step-by-step workflow of our RLEMMO approach. Firstly, we introduce the network architecture used in RLEMMO, as illustrated in Figure~\ref{fig:architecture}. The network parameters $\Theta$ consist of four parts, namely $\Theta := \{\Theta_{PE}, \Theta_{IE}, \Theta_{Actor}, \Theta_{Critic}\}$, where: 1) $\Theta_{PE} \in \mathbb{R}^{10 \times 32}$ and $\Theta_{IE} \in \mathbb{R}^{12 \times 32}$ represent the MLP embedders for population features $f_\text{pop}$ and individual features $f_\text{ind}$, respectively. 2) $\Theta_{Actor}$ denotes the strategy decoder (actor in RL) responsible for strategy sampling. It includes a $4$-heads self-attention block with a hidden embedding dimension of $64$ and an MLP layer with parameters in $\mathbb{R}^{64 \times 5}$. Note that we introduce the self-attention mechanism to enhance communication among individuals.  3) $\Theta_{Critic} \in \mathbb{R}^{64 \times 1}$ represents the MLP critic network used for estimating return values in RL. For the sake of clarity, we will omit the time stamp $t$ in the following derivation.

At each time step of the low-level optimization process, we start by aggregating the population features $f_\text{pop}(X)$ and individual features $f_\text{ind}(X)$ of all individuals in the current solution population $X$. Subsequently, we pass these two groups of features through separate embedders to obtain the corresponding feature embeddings:
\begin{equation}\label{eq:PEIE}
    PE = \Theta_{PE}(f_\text{pop}(X)),\quad IE=\Theta_{IE}(f_\text{ind}(X))
\end{equation}
We then obtain the decision embeddings for all individuals by concatenating their respective population feature embeddings ($PE$) and individual feature embeddings ($IE$):
\begin{equation}
    DE = hstack(\left[PE,IE\right])
\end{equation}
We feed the decision embeddings ($DE$) into the actor network and simultaneously sample the strategies ($A$) for all individuals from the soft distributions on its outputs, as defined in Section~\ref{Sec4.3}:
\begin{equation}\label{eq:actor}
    A \sim Softmax(\Theta_{Actor}(DE))
\end{equation}
The expectation of the return values ($G(X)$) is obtained by feeding the decision embeddings ($DE$) into the critic network:
\begin{equation}\label{eq:critic}
    G(X) =\Theta_{Critic}(DE)
\end{equation}
We present the pseudo code of the training workflow in our RLEMMO in Algorithm~\ref{alg1}. In this work, we utilize the Proximal Policy Optimization (PPO) method \cite{ppo} to meta-train the networks. Once the training process is complete, the trained RLEMMO model can be employed to solve unseen problems.  Note that during training, the strategy $A$ is sampled from the output distribution, while during inference, the strategy is selected greedily based on the maximum likelihood.
\begin{algorithm}[t]\small
	\caption{Pseudo code of RLEMMO} 
	\label{alg1} 
	\begin{algorithmic}
    \STATE \textbf{Input}: Training dataset $\Gamma$, Initial network parameters $\Theta^{(0)}$, Optimization horizon $T$, Learning epochs $MaxEpoch$, Learning rate $\eta$
    \STATE \textbf{Output}: Trained network parameters $\Theta$
        \FOR{$epoch = 0$ {\bfseries to} $MaxEpoch$}
        \FOR{each problem in $\Gamma$}
        \STATE  Initialize a solution population $X^{(0)}$;
        
        \FOR{$t = 0$ {\bfseries to} $T$}
        \STATE Get the state representation $\{f_\text{pop}(X^{(t)}),f_\text{ind}(X^{(t)})\}$.
        \STATE Sample actions $A^{(t)}$ following Eq.~(\ref{eq:PEIE}) $\sim$~(\ref{eq:actor}).
        \STATE Perform $A^{(t)}$ to update $X^{(t)}$ as $X^{(t+1)}$.
        \STATE Calculate $R_\text{clb}^{(t)}$ following Section~\ref{Sec4.4}.
		\STATE $\nabla_{\Theta,Critic} = \nabla_{\Theta}MSE(R_\text{clb}^{(t)}+G(X^{(t+1)}),G(X^{(t)}))$
        \STATE $\nabla_{\Theta,Actor} = \nabla_{\Theta}\log \mathbb{P}(A^{(t)}|\Theta)G(X^{(t)})$
        \STATE $\Theta^{(t+1)} = \Theta^{(t)} - \eta(\nabla_{\Theta,Actor}+\nabla_{\Theta,Critic})$
        \ENDFOR
        \ENDFOR
        \ENDFOR
	\end{algorithmic} 
\end{algorithm}
\section{Experimental results}
\begin{table}[t]
  \caption{Performance comparison of RLEMMO and the baselines, at the $10^{-4}$ accuracy level}
  \label{tab:compare}
  \resizebox{0.99\columnwidth}{!}{%
  \begin{tabular}{|c|c|cc|cc|cc|cc|cc|}
    \hline
     & \multirow{2}{*}{Functions}&\multicolumn{2}{c|}{RLEMMO} &\multicolumn{2}{c|}{VNCDE} &\multicolumn{2}{c|}{PNF-PSO} &\multicolumn{2}{c|}{EMO-MMO} &\multicolumn{2}{c|}{NBNC-PSO-ES} \\
     \cline{3-12}
    &&PR&SR&PR&SR&PR&SR&PR&SR&PR&SR\\
    \hline
    \multirow{14}*{\rotatebox{90}{Training}} &$F1$	&1	&1	&1	&1	&1	&1	&1	&1	&1	&1\\
    &$F3$	&1	&1	&1	&1	&1	&1	&1&	1&	1	&1\\
    &$F4$	&1	&1	&1	&1	&1	&1	&1	&1&	1	&1\\
    &$F6$	&\textbf{0.871}	&0.060	&0.577	&0	&0.652	&0	&0.827	&0.140&	0.692	&0\\
    &$F8$	&0.124	&0	&0.059	&0&	0.153	&0	&0.074	&0	&\textbf{0.165}	&0\\
    &$F9$	&0.211	&0	&0.177	&0	&0.073	&0	&\textbf{0.309}	&0	&0.083&	0\\
    &$F10$	&\textbf{1}	&1	&\textbf{1}&	1&	0.903&	0.280&	\textbf{1}	&1	&0.987	&0.840\\
    &$F12$	&\textbf{0.895}	&0.360&	0.740	&0	&0.698&	0.040&	0.705	&0	&0.778	&0.040\\
    &$F13$	&0.670&	0	&0.667	&0	&0.643	&0	&\textbf{0.800}	&0.100	&0.753	&0.020\\
    &$F17$	&0.263	&0	&0.265	&0	&0.145	&0	&0.143	&0	&\textbf{0.465}&	0\\
    &$F19$	&\textbf{0.341}&	0	&0.015&	0&	0	&0	&0.093	&0	&0.335&	0\\
    &$F20$	&0.155&0&	0&	0	&0	&0	&0.005&	0	&\textbf{0.225}&	0\\
    \cline{2-12}
    &Average&\textbf{ 0.628}&0.368&0.542&0.333&	0.522&0.277&	0.580&0.353&	0.624&0.325\\
    \cline{2-12}
    &Rank &\multicolumn{2}{c|}{\textbf{1.667}}&	\multicolumn{2}{c|}{2.833}&	\multicolumn{2}{c|}{3.500}&	\multicolumn{2}{c|}{2.250}&	\multicolumn{2}{c|}{1.917}\\
      

\cline{1-12}
    \multirow{10}*{\rotatebox{90}{Testing}}&$F2$	&1	&1	&1	&1	&1	&1	&1	&1	&1	&1\\
    &$F5$	&1	&1	&1	&1&	1&	1&	1&	1&	1&	1\\
    &$F7$	&0.703	&0	&0.644	&0	&0.324	&0	&\textbf{0.943}	&0.200	&0.420&	0\\
    &$F11$	&0.863	&0.320	&0.720&	0&	0.707	&0	&\textbf{0.963}	&0.820	&0.853&	0.180\\
    &$F14$	&0.667	&0	&0.667&	0	&0.623	&0	&0.640	&0&	\textbf{0.683}	&0\\
    &$F15$	&0.518&	0	&0.440&	0	&0.285	&0	&0.275	&0	&\textbf{0.585}	&0\\
    &$F16$	&0.640	&0&\textbf{	0.667}	&0	&0.190	&0&	0.377	&0	&0.653	&0\\
    &$F18$	&0.150&	0&	0.347&	0&	0.013&	0&	0.107	&0	&\textbf{0.410}	&0\\
    \cline{2-12}
    &Average	&0.693&0.290	&0.686&0.250&	0.518&0.250&	0.663&	0.378&\textbf{0.701}&0.273\\
    \cline{2-12}
    &Rank &\multicolumn{2}{c|}{2.000}&	\multicolumn{2}{c|}{2.125}&	\multicolumn{2}{c|}{3.875}&	\multicolumn{2}{c|}{2.625}&	\multicolumn{2}{c|}{\textbf{1.750}}\\

    \hline
  \end{tabular}
  }
\end{table}

\subsection{Experiment Setup}
\subsubsection{Dataset} We evaluate our RLEMMO on the well-known CEC 2013 MMOP benchmark \cite{cec2013}, which consists of $20$ multimodal problems with various dimensions and challenging landscape properties. And the number of optima varies from 1 to 216. For detailed definitions of the problems, please refer to Appendix \ref{appendixbenchmark}. We divide the $20$ problems into two groups: one for training and the other for testing. We carefully select problems with various dimensions for training, aiming to balance the distribution of problems with different dimension sizes in two datasets, to ensure the learning effectiveness and the generalization performance of our RLEMMO agent. The train-test split is detailed in Table~\ref{tab:cec2013benchmark}, Appendix \ref{appendixbenchmark}. 

\subsubsection{Metrics} Peak Ratio (PR) and Success Rate (SR) are two metrics commonly used to evaluate the performance of an MMOP algorithm \cite{cec2013}. Given a specified accuracy level (e.g., $10^{-4}$), we claim that an optimal of an MMOP is found when a solution is nearing that optimal with the accuracy. PR is calculated as the percentage of the number of found optima out of the total number of true optima. SR is calculated by defining a successful run as one in which all true optima have been found, and then calculating the ratio of the number of successful runs to the total number of runs. The formulas for calculating the two metrics are as follows:
\begin{equation}
PR = \frac{\sum_{i=1}^{NR}NPF_i}{NKP \cdot NR},\quad     SR = \frac{NSR}{NR}
\end{equation}
where $NR$ is the number of runs, $NPF_i$ is the number of found satisfactory solutions in the $i$-th run, $NKP$ is the number of known global optima, and $NSR$ is the number of successful runs.

\subsubsection{Baselines} We include several strong MMOP algorithms as baselines and compare our RLEMMO with them on the CEC 2013 benchmark using the PR/SR metrics. Specifically, we adopt VNCDE \cite{VNCDE} and NBNC-PSO-ES \cite{NBNC} as baselines that incorporate adaptive niching strategies to solve MMOP. We also include PNF-PSO \cite{EBSP} as a strong baseline for addressing MMOP through efficient spatial sampling. Additionally, EMO-MMO \cite{EMOMMO} serves as a baseline that solves MMOP by treating the global optima as independent objectives. 
We  note that although RLEA-SSC \cite{rlmmo} is an MetaBBO framework that incorporates RL at the meta level to control the selection operator during reproduction, we do not include it as a baseline due to its limitation in generalization. RLEA-SSC requires re-training on unseen problems, while our RLEMMO, once trained, can be directly used to solve unseen problems without the need for additional training.

\subsubsection{Settings} For our RLEMMO, the values of $F$ and $Cr$ for all actions in the designed action set are set to $0.5$ and $0.9$ respectively. The population size in the experiments is set to $100$. The value of $k$ for KNN is set to $4$, while the values of $minsample$ and $eps$ for DBSCAN are set to $3$ and $0.2$ respectively. We train RLEMMO for $60$ epochs, with a batch size of $4$ (i.e., initializing four populations for a sampled problem in a batch). The learning rate ($\eta$) starts at $5\times10^{-4}$ and linearly decays to $2\times10^{-4}$. The RL settings follow the original PPO paper \cite{ppo}. 
To ensure a fair and efficient evaluation, the population size ($NP$) is set to $100$, and the maximum function evaluations ($maxFEs$) is set to $50000$ for all baselines, including our RLEMMO.
The algorithmic configurations of the baselines follow their original papers. We evaluate each method for $50$ independent runs. The experiments were conducted on a computer with an Intel Gold 6254 CPU, an RTX 4090 GPU, and 64GB of RAM.

\begin{figure*}[t]
\centering
\subfigure[State Representation]{
\label{fig:abla-state}
\includegraphics[width=0.25\textwidth]{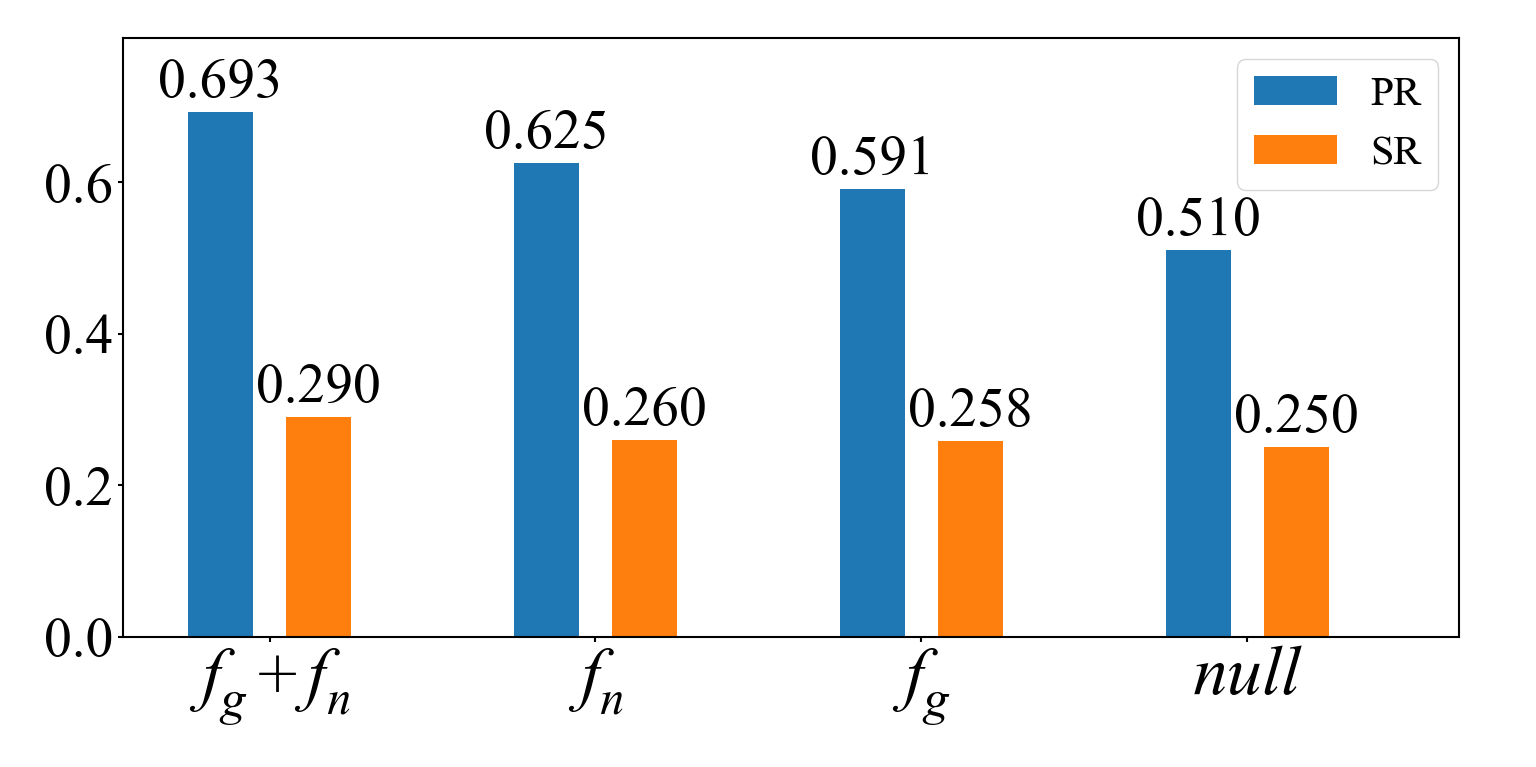}
}
\subfigure[Action Set]{
\label{fig:abla-action}
\includegraphics[width=0.25\textwidth]{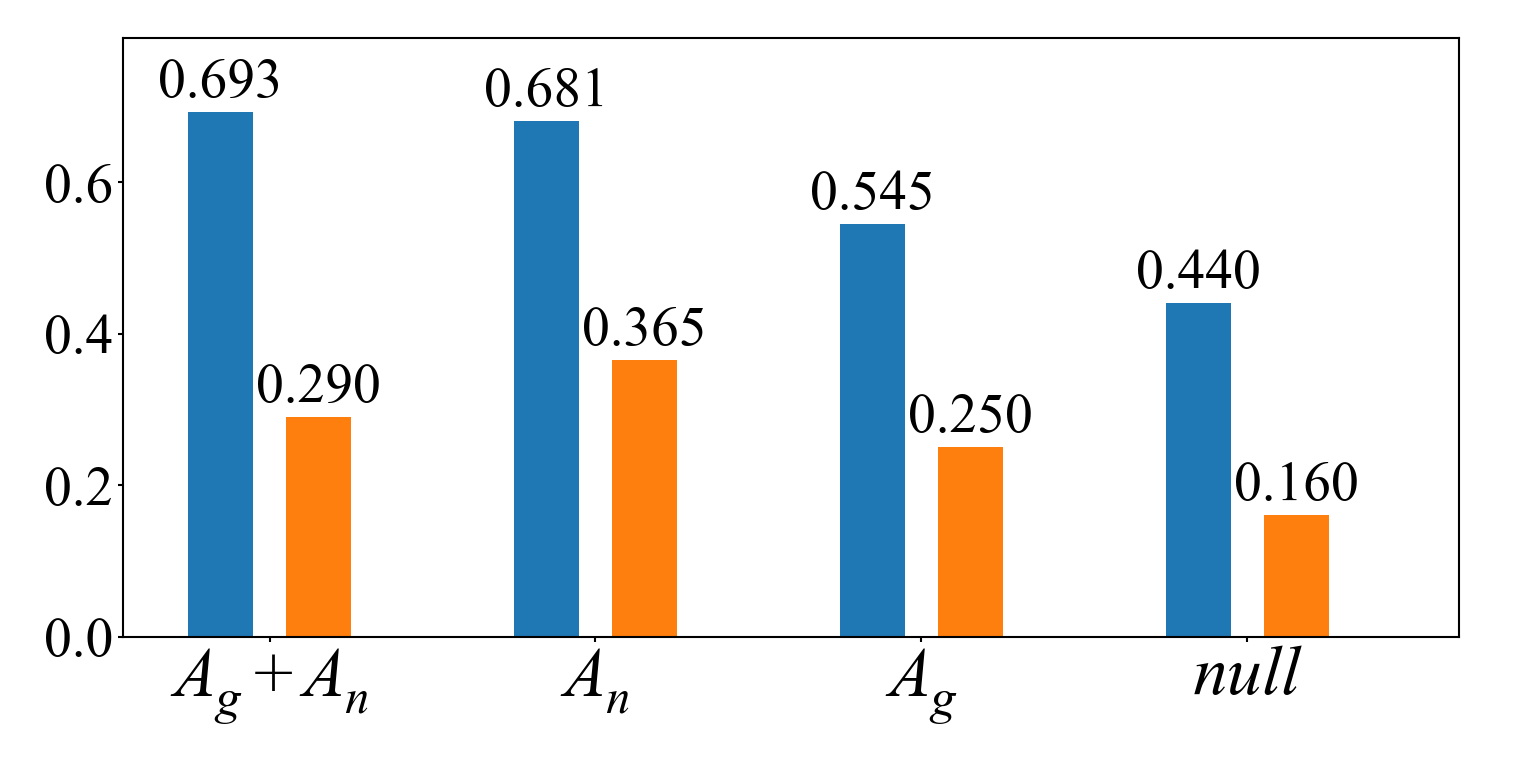}
}
\subfigure[Reward]{
\label{fig:abla-reward}
\includegraphics[width=0.25\textwidth]{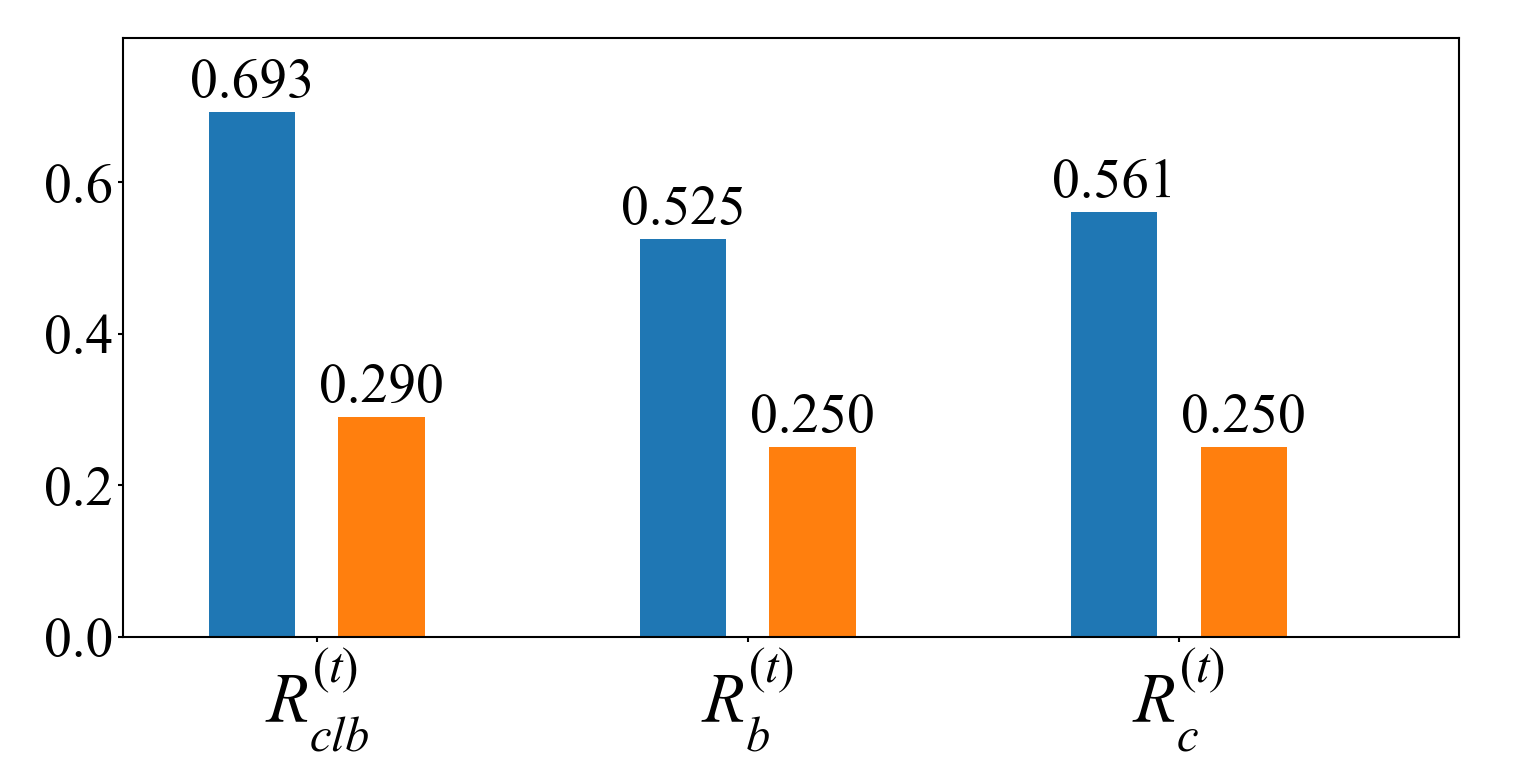}
}
\caption{Ablation studies: The average PR and SR of the ablation experiments on the testing dataset are compared at the accuracy level of $10^{-4}$. The results in sub-figures \ref{fig:abla-state} to \ref{fig:abla-reward} represent the ablation experiments on state features, action set, and reward mechanism, respectively.
}
\label{fig:abla}
\end{figure*}

\begin{table}[t]
  \caption{Average running time comparison of RLEMMO and the baselines}
  \label{tab:timecompare}
  \resizebox{0.95\columnwidth}{!}{%
  \begin{tabular}{|c|c|c|c|c|c|}
    \hline
     Algorithms&RLEMMO&VNCDE&PNF-PSO&EMO-MMO&NBNC-PSO-ES\\    
    \hline
    Running Time(s)&8.57&16.46&4.13&0.59&17.70\\    
    \hline
  \end{tabular}
  }
\end{table}
\subsection{Performance Analysis }
We report the optimization performance~(averaged over $50$ independent runs) and the runtime complexity of our RLEMMO and the considered baselines in Table~\ref{tab:compare} and Table~\ref{tab:timecompare}. The results demonstrate that:

\textbf{1.} Our RLEMMO demonstrates its effectiveness in solving MMOP. For the problems in the training set, RLEMMO achieves superior optimization performance compared to the baselines. On average, it locates more global optima (PR) and has more successful runs (SR). The optimization performance of RLEMMO on high-dimensional problems such as $F10$, $F12$, and $F19$ (which are challenging due to the large number of optima) demonstrates that the learned policy, which conditions on the up-to-date optimization status to flexibly switch the searching strategy, may be more effective in solving high-dimensional MMOP than the human-crafted rules used in the baselines.

\textbf{2.}The results on the test set highlight the potential generalization ability of RLEMMO. Once trained, RLEMMO is used to solve the problems in the test set without further fine-tuning. Although our algorithm does not rank the first in the comparison, we note that it demonstrates an acceptable level of generalization performance. The problems in the test set have different problem structures and landscape features. Despite this challenging setting, RLEMMO maintains competitive performance by achieving a similar PR performance to the best-performing NBNC-PSO-ES algorithm ($0.693$ versus $0.701$), while demonstrating a more robust searching performance considering the SR ($0.290$ versus $0.273$). Furthermore, we would like to clarify that the generalization ability of RLEMMO could be further improved by training on a problem distribution that includes various MMOPs, which is one of our future works.

\textbf{3.} RLEMMO demonstrates a more robust optimization performance across different landscape properties. We observe that some of the baselines encounter a significant performance drop across different problems, e.g., VNCDE from $F10$ to $F19$, and PNF-PSO from $F14$ to $F18$. Note that $F10$ and $F19$ have different problem structures, while $F14$ and $F18$ share the same structure but differ in dimensions. In contrast, RLEMMO exhibits a relatively small performance drop in these situations. This may reveal that the human-crafted rules in some baselines would fail to generalize across certain problem classes. RLEMMO, however, meta-learns to adaptively alter its searching pattern for fitting unseen problems. 

\textbf{4.} From the results in Table~\ref{tab:timecompare}, it is evident that RLEMMO exhibits an acceptable running complexity compared to other algorithms.

\subsection{Ablation Study }
To validate the effectiveness of the core designs in our work, we conducted several ablation studies from different aspects, including the state representation, the action set, and the design of the reward. Figure \ref{fig:abla} shows the average PR (blue bar) and SR (orange bar) on the test set at an accuracy level of $10^{-4}$.

\subsubsection{State Representation}
To quantify the importance of $f_\text{pop}$, we conducted ablation experiments by removing $f_{g}$, $f_{n}$, or both from our $f_\text{pop}$ and trained three models under these settings. In Figure~\ref{fig:abla-state}, we present the performance of these three models (denoted as `$f_n$', `$f_g$' and `$null$', respectively ) as well as the original model (denoted as  `$f_g+f_n$'). The results indicate that: 1) Our proposed neighbor-aware population feature $f_\text{pop}$ plays a key role in RLEMMO, enabling the meta-level RL agent to be fully informed about both the global and local landscape status. 2) The local information $f_n$ may contribute relatively more than the global distributional feature, as it biases the optimization status to provide fine-grained state representations.


\subsubsection{Action}
To quantify the importance of maintaining diverse searching strategies~($A_1$ to $A_5$) for the final optimization performance of RLEMMO, we conducted ablation experiments by removing global strategies~($A_g=\{A_4,A_5\}$), niching strategies~($A_n=\{A_2,A_3\}$), or both from the action set $A$ and trained three RLEMMO models under these three settings. Figure~\ref{fig:abla-action} presents the performance of these three models (denoted as  `$A_n$',  `$A_g$', and  `$null$', respectively) as well as the original model (denoted as  `$A_g+A_n$'). The results demonstrate that: 1) The action set containing both global strategies  $A_g$ and niching strategies  $A_n$ contributes to locating more optima by promoting optimization and maintaining diversity. 2) The niching strategies $A_n$ are relatively more important than the global strategies $A_g$ in solving MMOPs, underscoring the necessity of maintaining population diversity.

\subsubsection{Reward}
The reward mechanism in RLEMMO encourages two aspects of optimization: increasing the number of clusters and improving the object values within those clusters. This approach motivates the policy to maintain population diversity while striving for optimal solutions. To assess the importance of these two aspects, we have designed two new rewards that focus on each objective individually. The first reward, denoted as $R_b^{(t)}$, is calculated using the best objective value in the population at generation $t$, represented as $f^{*,(t)}$, encouraging the optimization of quality. The second reward, denoted as $R_c^{(t)}$, is computed based on the number of clusters formed by the population at generation $t$, represented as $Count(cluster^{(t)})$, considering the diversity of the population. The formulas for these rewards are as follows:
\begin{equation}
    R_b^{(t)} = f^{*,(t)},
    R_c^{(t)} = Count(cluster^{(t)})
\end{equation}
The results of the experiment with two new rewards (denoted as  `$R_b^{(t)}$' and  `$R_c^{(t)}$') and the original experiment (denoted as `$R_{clb}^{(t)}$') are shown in Figure \ref{fig:abla-reward}. The results indicate that: 1) The reward related to both diversity and quality is crucial for locating multiple optima, as it encourages the population to maintain diversity while optimizing for optimal solutions. 2) The reward related to diversity has a relatively greater influence than the one related to quality, enhancing the ability to discover more optimal solutions within a single run.


\section{Conclusion}

In this paper, we propose RLEMMO, the first generalizable Meta-Black-Box Optimization framework, for solving multimodal optimization problems using reinforcement learning. By incorporating a fitness landscape analysis based state representation informing the accurate local and global optimization status, the RL agent can be efficiently meta-learned through a novel clustering-based reward scheme on a problem distribution. Once trained, RLEMMO can be directly used to solve unseen problems, achieving competitive optimization performance against several strong MMOP solvers, both in quality and diversity. As a pioneering work, RLEMMO has certain improvement space, of which enriching the strategy diversity and a more fine-grained state representation are our immediate future works.

\begin{acks}
This work was supported in part by the National Natural Science Foundation of China under Grant 62276100, in part by the Guangdong Natural Science Funds for Distinguished Young Scholars under Grant 2022B1515020049, in part by the Guangdong Regional Joint Funds for Basic and Applied Research under Grant 2021B1515120078, and in part by the TCL Young Scholars Program.
\end{acks}

\bibliographystyle{ACM-Reference-Format}
\bibliography{sample-base}

\appendix
\section{Per-feature Exegesis}~\label{appendixA}
The state representation for the optimization status in RLEMMO consists of a total of $22$ features. It contains two aspects: $f_{ind}$, which describes the environment of each individual, and $f_{pop} = \{f_g, f_n\}$, which describes the landscape of the entire population and the neighborhood around each individual. Next, we will explain each feature in three parts respectively.
\begin{itemize}
\item For $f_g$, we introduce five simple yet informative features to provide the RL agent with a global view of the population. $f_1 = \text{mean}_{i,j}(\frac{\| X_i^{(t)} - X_j^{(t)} \|_2}{diameter})$ describes the distance between each pair in the population, and $f_2 = \text{std}_i(\frac{Obj(X_i^{(t)})}{Obj_\text{max}})$ describes the standard deviation of objective values. These two features represent the global distribution and convergence status.  $f_{3}=\frac{(T-t)}{T}$ provides the remaining portion of generations during the optimization progress. And $f_4=\frac{st}{T}$ is the stagnation of the global best solution, reflecting the potential for exploitation. $f_{5}=\text{mean}_i(\frac{Obj(X_i^{(t)})}{Obj_\text{max}})$ is the average objective value of the population, representing the optimization level of the population. Note that since $f_g$ is a global state, so it is the same for all individuals in the population. From these features, the agent can learn about the knowledge for global optimization.
\item $f_n$ for each individual $X_i$ is composed of the information within its neighborhood $\Phi_i$, which is constructed by  KNN \cite{KNN}, providing the agent with informative local information around each individual. $f_6 = \text{mean}_{j,k\in \Phi_i^{(t)}}(\frac{{\|X_j^{(t)} - X_k^{(t)}\|}_2}{diameter})$ profiles the distance of each pair within $\Phi_i$, and the standard deviation  of objective values within $\Phi_i$  is calculated as $f_7 = \text{std}_{j\in \Phi_i^{(t)}}(\frac{Obj(X_j^{(t)})}{Obj_\text{max}})$ . These two feature represent the local distribution status around each individual. $f_8=\frac{st(\Phi_i)}{T}$ is the stagnation of the best objective value within $\Phi_i$, reflecting the potential for local exploitation.  $f_9 = \text{mean}_{j\in \Phi_i^{(t)}}(\frac{Obj(X_j^{(t)})}{Obj_\text{max}})$  describes the local optimization progress by the average objective value within $\Phi_i$. An additional feature $f_{10}=\frac{\text{rank}({\Phi_i^{(t)}})}{(NP-1)}$ is introduced to profile the performance of $\Phi_i$ among the whole population by ranking all neighborhoods based on the best fitness in each neighborhood. These features provide the knowledge about landscape for decision about local search.
\item $f_{ind}$ describes the information about each individual itself, denoted from $f_{11}$ to $f_{22}$. Firstly, we utilize features related to the distance and objective value difference from the best solutions to coordinate the progress in exploitation. To be specific, $f_{11}=\frac{{\|X_i^{(t)} - X^{*,{(t)}}\|}_2}{diameter}$ and $f_{14}=\frac{Obj(X_i^{(t)}) - Obj(X^{*,{(t)}})}{Obj_\text{max}}$ are calculated by the distance and the difference of objective value between $X_i$ and the best solution in the current population, while $f_{12}=\frac{{\|X_i^{(t)} - X^*\|}_2}{diameter}$ and $f_{13}=\frac{Obj(X_i^{(t)}) - Obj(X^*)}{Obj_\text{max}}$ provide information between $X_i$ and  the historical best solution. These four features guide for the global exploitation for $X_i$. $f_{15}=\frac{{\|X_i^{(t)} - \Phi_i^{*}\|}_2}{diameter}$ and $f_{16}=\frac{Obj(X_i^{(t)}) - Obj(\Phi_i^{*})}{Obj_\text{max}}$ inform the relationship between $X_i$ and the best solution in its neighborhood $\Phi_i$, guiding the local exploitation for $X_i$.  Secondly, to provide the agent with a perspective on exploration, we utilize the information 
 about the pairs within the population. Here, $f_{19}=\text{mean}_{j \in \Phi_i^{(t)}}(\frac{Obj(X_i^{(t)}) - Obj(X_j^{(t)})}{Obj_\text{max}})$ and $f_{20}=\text{mean}_{j \in \Phi_i^{(t)}}(\frac{{\|X_i^{(t)} - X_j^{(t)}\|}_2}{diameter})$ are the average difference of objective values and the average distances between $X_i$ and each neighbors within $\Phi_i$, which provides the local environment within $\Phi_i$. $f_{21}=\text{mean}_j(\frac{Obj(X_i^{(t)})- Obj(X_j^{(t)})}{Obj_\text{max}})$ and $f_{22}=\text{mean}_j(\frac{{\|X_i^{(t)} - X_j^{(t)}\|}_2}{diameter})$ describe the relationship between $X_i$ and other individuals in the entire population, reflecting the global environment about $X_i$. These aforementioned features provide the necessary knowledge for the agent to balance exploitation and exploration, whether it is at a local or global level. Also, the stagnation and absolute performance of $X_i$ are included as $f_{17}=\frac{st(i)}{T}$ and $f_{18}=\frac{Obj(X_i^{(t)})}{Obj_\text{max}}$.
\end{itemize}

The aforementioned $22$ features provide comprehensive knowledge of each individual about current landscape and evolution-path, assisting the agent to select appropriate search strategies to match the optimization status.

\section{Benchmark}~\label{appendixbenchmark}

\begin{table}[t]
  \caption{Parameters of problems in CEC2013 benchmark}
  \label{tab:cec2013benchmark}
  \resizebox{0.95\columnwidth}{!}{%
  \begin{tabular}{|c|c|c|c|c|c|}
    \hline
     &Problem Number&Function&r&Peak height&No. global optima\\    
    \hline
    \multirow{12}*{\rotatebox{90}{Training}}&$F1$&$F_{1}$(1D)&0.01&200.0&2\\
    &$F3$&$F_{3}$(1D)&0.01 & 1.0&1\\
    &$F4$&$F_{4}$(2D)&0.01 & 200.0&4\\
    &$F6$&$F_{6}$(2D)&0.5 & 186.731&18\\
    &$F8$&$F_{6}$(3D)& 0.5& 2709.0935&81\\
    &$F9$&$F_{7}$(3D)& 0.2& 1.0&216\\
    &$F10$&$F_{8}$(2D)& 0.01& -2.0&12\\
    &$F12$&$F_{10}$(2D)& 0.01& 0&8\\
    &$F13$&$F_{11}$(2D)& 0.01& 0&6\\
    &$F17$&$F_{12}$(5D)&0.01 & 0&8\\
    &$F19$&$F_{12}$(10D)&0.01 & 0&8\\
    &$F20$&$F_{12}$(20D)& 0.01& 0&8\\
    \hline
    \multirow{8}*{\rotatebox{90}{Testing}} &$F2$&$F_{2}$(1D)&0.01 &1.0 &5\\
    &$F5$&$F_{5}$(2D)&0.5 & 1.03163&2\\
    &$F7$&$F_{7}$(2D)& 0.2& 1.0&36\\
    &$F11$&$F_{9}$(2D)&0.01 & 0&6\\
    &$F14$&$F_{11}$(3D)& 0.01& 0&6\\
    &$F15$&$F_{12}$(3D)&0.01 & 0&8\\
    &$F16$&$F_{11}$(5D)& 0.01& 0&6\\
    &$F18$&$F_{11}$(10D)&0.01 & 0&6\\
    \hline
  \end{tabular}
  }
\end{table}

The CEC2013 MMOP Benchmark consists of 20 problems, each generated by a specified function and dimension size. There are $12$ functions used to construct the problem set, with the first $8$ functions being simple and well-known functions, while the remaining problems are complex composition functions. Each composition function is formulated based on the following equation:
\begin{equation}
    CF_j(\vec x)=\sum_{i=1}^n w_i(\hat{f}_i((\vec{x}-\vec{\sigma_i})/\lambda _i \cdot M_i) + bias_i) + f_{bias} ^j
\end{equation}
where $n$ is the number of basic functions, $\vec f_i$ denotes the normalization of the $i$-th basic function, and $w_i$ is the corresponding weight. $\vec \sigma_i$, $M_i$ and $\lambda_i$ are the new shifted optimum, linear transformation matrix and a parameter used to stretch or compress for each $\hat f_i$, respectively. Additionally, $bias_i$ and $f_{bias}^j$ are two bias parameters. The pool of basic functions used to construct composition functions is listed as follows:
\begin{enumerate}
\item Sphere function: 
\begin{equation}
    f_S(\vec{x}) = \sum_{i=1}^D x_i^2
\end{equation}
\item Grienwank's function: 
\begin{equation}
    f_G(\vec{x}) = \sum_{i=1}^D \frac{x_i^2}{4000} - \prod_{i=1}^D cos(\frac{x_i}{\sqrt{i}})+1
\end{equation}
\item Rastrigin's function: 
\begin{equation}
    f_R(\vec{x})=\sum_{i=1}^D (x_i^2- 10cos(2\pi x_i)+10)
\end{equation}
\item Weierstrass function: 
\begin{equation}
\begin{aligned}
    f_W(\vec{x}) = \sum_{i=1}^D(\sum_{k=0}^{kmax}\alpha ^k cos(2\pi \beta ^k(x_i+0.5)))\\
    -D\sum_{k=0}^{kmax}\alpha ^k cos(2\pi \beta ^k(0.5))
\end{aligned}
\end{equation} 
where $\alpha = 0.5, \beta = 3, kmax = 20$. 
\item Expanded Griewank's plus Rosenbrock's function(EF8F2):
\begin{equation}
    F8(\vec x) = \sum_{i=1}^D \frac{x_i^2}{4000} - \prod_{i=1}^D cos(\frac{x_i}{\sqrt{i}})+1
\end{equation}
\begin{equation}
F2(\vec x) = \sum_{i=1}^{D-1}(100(x_i^2 - x_{i+1})^2 + (x_i -1)^2)
\end{equation}
\begin{equation}
\begin{aligned}
   & EF8F2(\vec x) = F8F2(x_1,x_2,...,x_D) \\
   &=F8(F2(x_1,x_2))+F8(F2(x_2,x_3))+...\\
   &+F8(F2(x_{D-1},x_D))+F8(F2(x_D,x_1))
\end{aligned}
\end{equation}

\end{enumerate}

Using these basic functions, several composition functions have been constructed for the benchmark, whose brief information is listed below, following the formulations of the simple functions in the benchmark. More details about the parameters and characters of each function can be referred in~\cite{cec2013}. 

\begin{itemize}
\item $F_{1}$ : Five-Uneven-Peak Trap
\begin{equation}
     F_1(x)=\left\{
\begin{array}{rcl}
80(2.5-x)       &      & {\text{for } 0 \leq x <    2.5}\\
64(x-2.5)     &      & {\text{for } 2.5 \leq x < 5.0}\\
64(7.5-x)     &      & {\text{for } 5.0 \leq x < 7.5}\\
28(x-7.5)       &      & {\text{for } 7.5 \leq x < 12.5}\\
28(17.5 - x)       &      & {\text{for } 12.5 \leq x  <  17.5  }\\
32(x-17.5)     &      & {\text{for } 17.5 \leq x < 22.5}\\
32(27.5-x)     &      & {\text{for } 22.5 \leq x < 27.5}\\
80(x-27.5)       &      & {\text{for } 27.5 \leq x  \leq  30}
\end{array} \right. 
\end{equation}
\item $F_{2}$ : Equal Maxima
\begin{equation}
    F_2(x) = sin^6(5\pi x)
\end{equation}
\item $F_{3}$ : Uneven Decreasing Maxima
\begin{equation}
    F_3(x) = exp(-2\log(2)(\frac{x-0.08}{0.854})^2)sin^6(5\pi (x^{3/4} - 0.05))
\end{equation}
\item $F_{4}$ : Himmelblau
\begin{equation}
    F_4(x,y) = 200-(x^2+y-11)^2 - (x+y^2-7)^2
\end{equation}
\item $F_{5}$ :  Six-Hump Camel Back
\begin{equation}
    F_5(x,y) = -4[(4-2.1x^2+\frac{x^4}{3})x^2 +xy+(4y^2-4)y^2]
\end{equation}
\item $F_{6}$ : Shubert
\begin{equation}
    F_6(\vec{x}) = - \prod_{i=1}^D\sum_{j=1}^5jcos[(j+1)x_i+j]
\end{equation}
\item $F_{7}$ :  Vincent
\begin{equation}
    F_7(\vec{x}) = \frac{1}{D}\sum_{i=1}^D sin(10\log(x_i))
\end{equation}
\item $F_{8}$ :  Modified Rastrigin - All Global Optima
\begin{equation}
    F_8(\vec{x}) = -\sum_{i=1}^D(10+9cos(2\pi k_ix_i))
\end{equation}
\item $F_{9}$ : Composition Function 1
\begin{itemize}
\item $f_1 - f_2$ : Grienwank's function, 
\item $f_3 - f_4$ : Weierstrass function,
\item  $f_5-f_6$ : Sphere function.   
\item  $\sigma_i = 1, \forall i \in \{1,2,...,n\}$
\item  $\vec{\lambda} = [1,1,8,8,1/5,1/5]$
\item $M_i$ are identity matrices $\forall i \in \{1,2,...,n\}$
\end{itemize}
\item $F_{10}$ : Composition Function 2
\begin{itemize}
    \item $f_1 - f_2$ : Rastrigin's function, 
\item $f_3 - f_4$ : Weierstrass function,
\item  $f_5-f_6$ : Griewank's function,
\item $f_7-F_8$ : Sphere function.
\item $\sigma_i = 1, \forall i \in \{1,2,...,n\}$
\item  $\vec{\lambda} = [1,1,10,10,1/10,1/10,1/7,1/7]$
\item $M_i$ are identity matrices $\forall i \in \{1,2,...,n\}$
\end{itemize}
\item $F_{11}$ : Composition Function 3
\begin{itemize}
\item $f_1 - f_2$ : EF8F2 function, 
\item $f_3 - f_4$ : Weierstrass function,
\item  $f_5-f_6$ : Griewank's function.    
\item $\sigma_i = [1,1,2,2,2,2]$
\item  $\vec{\lambda} = [1/4,1/10,2,1,2,5]$
\item $M_i$ are different linear transformation matrices with condition number one.
\end{itemize}
\item  $F_{12}$ :  Composition Function 4
\begin{itemize}
\item $f_1 - f_2$ : Rastrigin's function, 
\item $f_3 - f_4$ : EF8F2 function,
\item  $f_5-f_6$ : Weierstrass function,
\item  $f_7-f_8$ : Griewank's function.
\item $\sigma_i = [1,1,1,1,1,2,2,2]$
\item  $\vec{\lambda} = [4,1,4,1,1/10,1/5,1/10,1/40]$
\item $M_i$ are different linear transformation matrices with condition number one.
\end{itemize}
\end{itemize}

Having these functions, $20$ problems in the benchmark can be constructed with assigned dimension size. The parameters for each problem are shown in Table~\ref{tab:cec2013benchmark}. Note that we divide the problems into two groups according to the training and testing settings used in our experiments on RLEMMO.

\end{document}